\documentclass[runningheads]{llncs}
\usepackage[T1]{fontenc}
\usepackage{accvabbrv}
\usepackage{graphicx}
\usepackage{booktabs}
\usepackage{color}
\PassOptionsToPackage{prologue,dvipsnames}{xcolor}
\usepackage[cmyk]{xcolor}
\usepackage{colortbl}
\usepackage{multirow}
\usepackage{float}
\usepackage{hyperref}
\usepackage{amsmath}
\usepackage{amssymb}
\begin{document}
\title{Wavelet-based Mamba with Fourier Adjustment for Low-light Image Enhancement} 

\titlerunning{WalMaFa}

\author{Junhao Tan$^{\dag}$\inst{1} \and
Songwen Pei$^{\dag*}$\inst{1} \and
Wei Qin\inst{1} \and
Bo Fu\inst{2} \and
Ximing Li\inst{3} \and
Libo Huang\inst{4}
{\\ \small{{$^{\dag}$}Contribute equally~~${*}$Corresponding author. Email address: swpei@usst.edu.cn}}}
\authorrunning{J. Tan et al.}
\institute{School of Optical-Electrical and Computer Engineering, University of Shanghai for
Science and Technology, Shanghai, 200093, China\\
\email{223330941@st.usst.edu.cn}, \email{swpei@usst.edu.cn}, \email{201440056@st.usst.edu.cn}
\and
School of computer science and artificial intelligence, Liaoning Normal University, Liaoning, 116081, China
\and
College of computer science and technology, Jilin University, Jilin, 134000, China
\and
School of Computer, National University of Dense Technology, Changsha, 410073, China
}
\maketitle
\begin{abstract}
Frequency information (\eg, Discrete Wavelet Transform and Fast Fourier Transform) has been widely applied to solve the issue of Low-Light Image Enhancement (LLIE). However, existing frequency-based models primarily operate in the simple wavelet or Fourier space of images, which lacks utilization of valid global and local information in each space. We found that wavelet frequency information is more sensitive to global brightness due to its low-frequency component while Fourier frequency information is more sensitive to local details due to its phase component. In order to achieve superior preliminary brightness enhancement by optimally integrating spatial channel information with low-frequency components in the wavelet transform, we introduce channel-wise Mamba, which compensates for the long-range dependencies of CNNs and has lower complexity compared to Diffusion and Transformer models. So in this work, we propose a novel Wavelet-based Mamba with Fourier Adjustment model called \textbf{WalMaFa}, consisting of a Wavelet-based Mamba Block (WMB) and a Fast Fourier Adjustment Block (FFAB). We employ an Encoder-Latent-Decoder structure to accomplish the end-to-end transformation. Specifically, WMB is adopted in the Encoder and Decoder to enhance global brightness while FFAB is adopted in the Latent to fine-tune local texture details and alleviate ambiguity. Extensive experiments demonstrate that our proposed WalMaFa achieves state-of-the-art performance with fewer computational resources and faster speed. Code is now available at: \href{https://github.com/mcpaulgeorge/WalMaFa}{https://github.com/mcpaulgeorge/WalMaFa}.

\end{abstract}

\begin{figure}[tb]
\centering
\includegraphics[width=\linewidth]{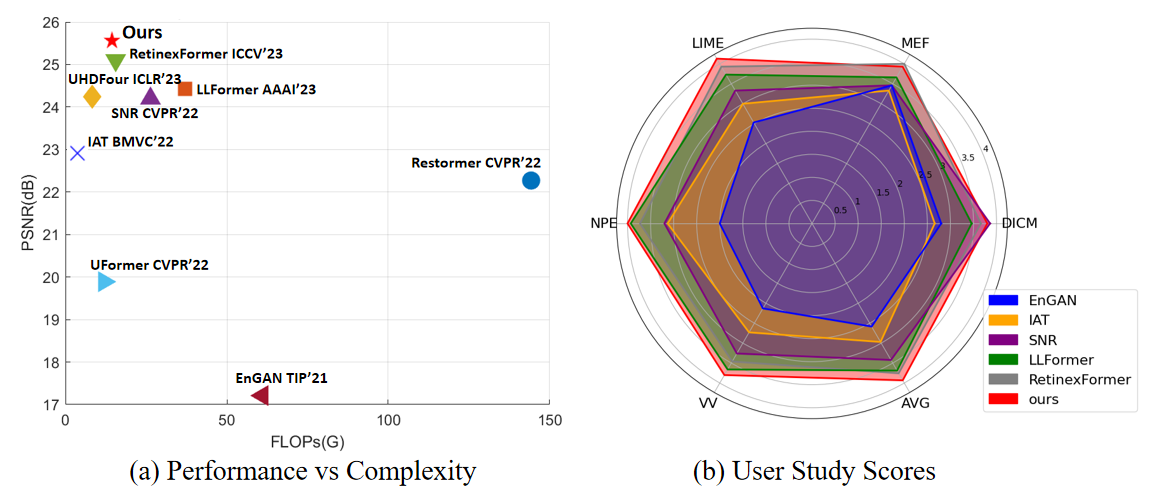}\\
\caption{WalMaFa consistently achieves relatively better performance and less computing complexity
on LOL-v2-syn dataset. WalMaFa also stands out in human-perceived user ratings with 15 participants.}\label{fig:1} 
\end{figure}
\section{Introduction}
\label{sec:intro}
Low-light images suffer from multiple visual degradations, including low resolution with intensive noises, brightness degradations and colour distortions. These issues significantly interfere with human perception of image information and affect various visual downstream tasks (\eg, medical image segmentation\cite{10.1007/978-981-99-7025-4_42} and autonomous driving\cite{9874070}). Low-light issues can be improved by upgrading the hardware of the capturing equipment, but the costs are very high. Thus, Low-light Image Enhancement (LLIE), which aims to recover hidden global or local degradations, is considered an active approach to post-process low-light images in computer vision.

The learning-based models have been proposed with great performance, including CNN-based\cite{DeepUPE,ZeroDCE} methods, Transformer-based\cite{Uformer,Restormer,LLFormer,SNR} methods and Retinex-based methods\cite{Retinexnet,retinexformer}. However, most of them are based on spatial features in raw space rather than considering frequency information. Furthermore, CNN lacks long-range dependency, leading to poor global enhancement, while Transformers require more computational resources, resulting in slower inference speed and higher resource consumption.
\begin{figure}[tb]
\centering
\includegraphics[width=\linewidth]{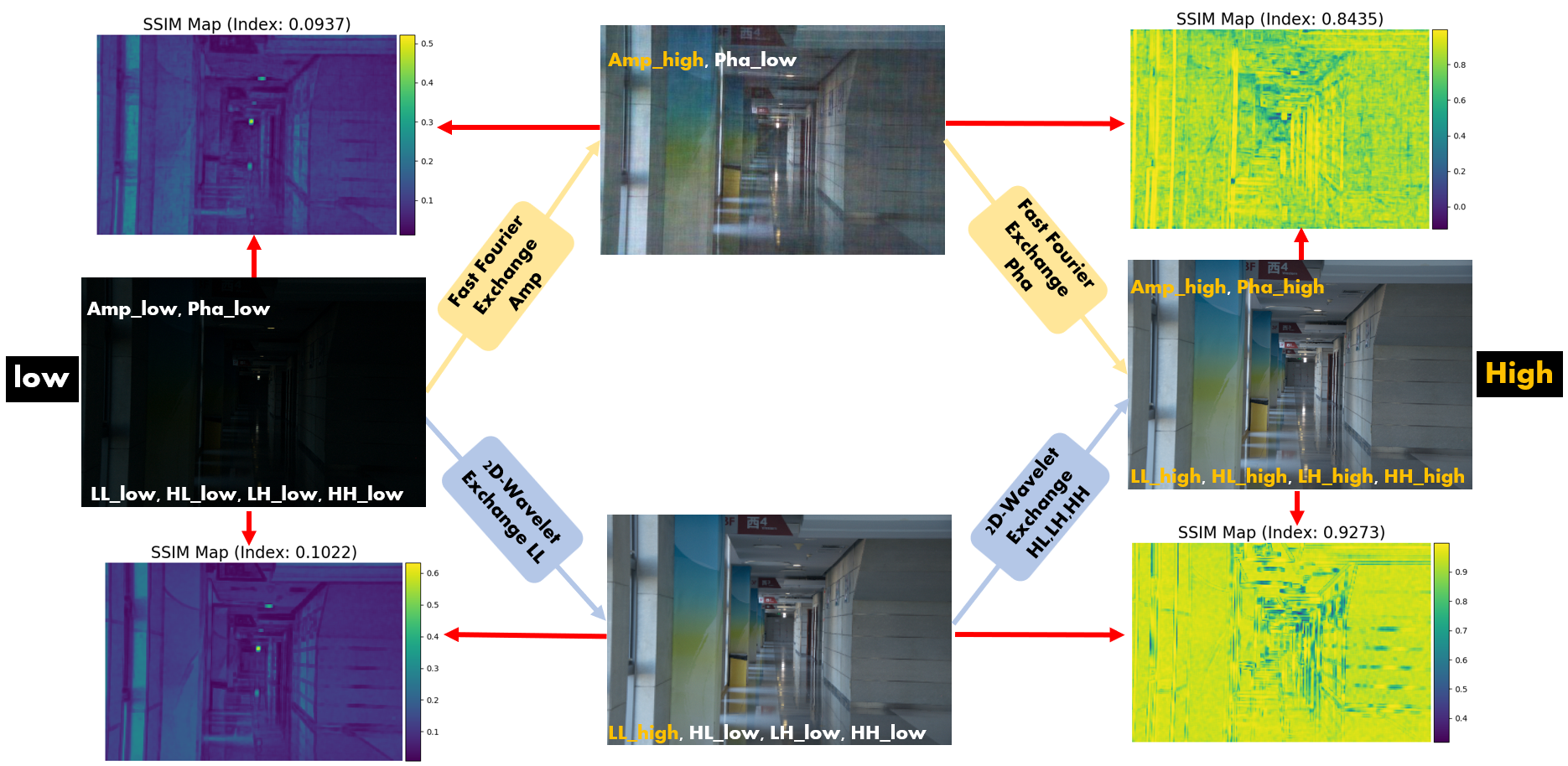}\\
\caption{The motivation of WalMaFa. The SSIM error map between the output after switching different components and the low/high image shows that the low-frequency LL component of 2D Discrete Wavelet Transform is more sensitive to global color brightness, while the phase component of Fast Fourier Transform is more sensitive to local texture detail information.}\label{fig:2} 
\end{figure}

Recently, some methods\cite{IGAWN} have been exploring the wavelet frequency information for LLIE. They combine wavelet frequency information with spatial information via Transformer, which reveals that the low-frequency component encodes global information and the high-frequency component encodes local details. Meanwhile, some methods\cite{UHDFourICLR2023} explore the Fourier frequency information. They combine Fourier frequency information with spatial information through CNN, which proves that the amplitude frequency component represents the global brightness, while the phase frequency component represents the local texture details.

Obviously, both wavelet and Fourier transforms exhibit frequency components with similar properties. So, which of their components is better? Our motivation is a good answer to this question, as illustrated in Fig. \ref{fig:2}. Firstly, given two images (\ie, a low-light image and a normal-high image), we swap their low-frequency LL component of the wavelet transform and the amplitude component of the Fourier transform while preserving the remaining components. Then, we swap the remaining components on top of the previous swapping. To visualize the enhancements after swapping components, we calculated the SSIM map between the swapped image and the low/high image. The results show that the image after swapping the LL component is brighter and clearer than the image after swapping the amplitude component. Numerically, the image after swapping the LL component is also 0.08dB higher (0.9273 vs. 0.8435) in terms of SSIM value compared to the high image. However, the image after swapping the phase component makes up for the previous 0.08dB difference between LL and amplitude, which reveals that phase component in Fourier transform can capture more local details than the high-frequency components \{LH, HL, HH\} in the wavelet transform.

Based on the above motivation, in this work, we propose a novel Encoder-Latent-Decoder method called Wavelet-based Mamba with Fourier Adjustment (WalMaFa), which consists of Wavelet-base Mamba Blocks (WMB) and Fast Fourier Adjustment Blocks (FFAB). WMB is adopted in the encoder and decoder to enhance global brightness, while FFAB is utilized in the latent to adjust local texture details and reduce ambiguity. To enhance global brightness specifically, we introduce channel-wise Mamba to extract low-frequency channel information of WMB. This approach leverages Mamba’s ability for linear analysis of long-distance sequences and offers lower computational complexity compared to Transformer. In this way, our model takes into account both global brightness and local details.

Comprehensive experiments shown in Fig. \ref{fig:1}(a) prove the excellent performance and fewer computing complexity of WalMaFa. Besides, we conduct user human-perceived study with 15 participants, as shown in Fig. \ref{fig:1}(b) to verify the superior perceptual quality of WalMaFa.

The main contributions of this paper are summarized as:

$\bullet$ We propose a novel Wavelet-based Mamba Block (WMB) to capture more global brightness information by combining spatial channel information of Channel Mamba with low-frequency information of the wavelet transform.

$\bullet$ We propose a novel Fast Fourier Adjustment Block (FFAB). On top of the global brightening of WMB, local texture details are adjusted thanks to Fourier's phase component enhancement, resulting in a smoother and clearer result.
 
$\bullet$ Comprehensive experiments on classic Low-light Image Enhancement benchmarks demonstrate superior performance and complexity. Besides, we conduct a user study with 15 participants to verify the superior perceptual quality of WalMaFa.

\section{Related Works}

\noindent\textbf{Low-light Image Enhancement.} With the rapid development of deep learning, many deep-learning\cite{ZeroDCE,SNR,IPT,Restormer,IAT} methods have been proposed to solve the Low-light Image Enhancement. Guo \etal~\cite{ZeroDCE} proposed a zero-reference deep curve estimation method to enhance unpaired low-light images. Cui \etal~\cite{IAT} proposed a lightweight Transformer IAT with only 0.09M parameters. Xu \etal~\cite{SNR} combined SNR map with Transformer to LLIE. Retinex-based methods \cite{retinexformer,Retinexnet} introduced the illumination map of low-light images. However, the above methods only operated in raw spatial space, limiting their ability to leverage the fusion of frequency domain information with spatial information.

To make full use of frequency domain information, Xu \etal~\cite{IGAWN} proposed an attentive wavelet network utilising wavelet frequency information and spatial information of attention. Li \etal~\cite{UHDFourICLR2023} embedded Fourier frequencies into the network. Although they discovered the role of frequency components in the wavelet and Fourier transform, they did not compare the corresponding components.

Unlike the above methods, we found that the low-frequency LL component of the wavelet transform is more sensitive to global color brightness than the amplitude component of the Fourier transform, while the phase component of the Fourier transform is more sensitive to local texture detail information than the high-frequency component of the wavelet transform. So, our WalMaFa employs an Encoder-Latent-Decoder-structured network, where wavelet frequency is applied to Encoder and Decoder for preliminary brightness enhancement, while Fourier frequency is applied to the Latent for detail adjustment.

\noindent\textbf{State Space Models.} State Space Models (SSMs) have recently received a great deal of attention. SSMs draw inspiration from state space models in control systems and aim to address long-range dependency issues, as expressed in LSSL\cite{LSSL}. To further overcome the huge complexity of LSSL, S4\cite{S4} was proposed as a alternative to CNNs and Transformers for capturing long-range dependencies.

More recently, a generic language model backbone Mamba\cite{mamba} has been proposed as a selective state space model that enables context-dependent reasoning while scaling linearly in sequence length. Inspired by this research, our work leverages Mamba’s capability for linear analysis of long-distance sequences to process the low-frequency LL component of the wavelet transform in the channel dimension, which enriches the fusion of spatial brightness information and low-frequency brightness information. 

\begin{figure}[t]
\centering
\includegraphics[height=11.0cm]{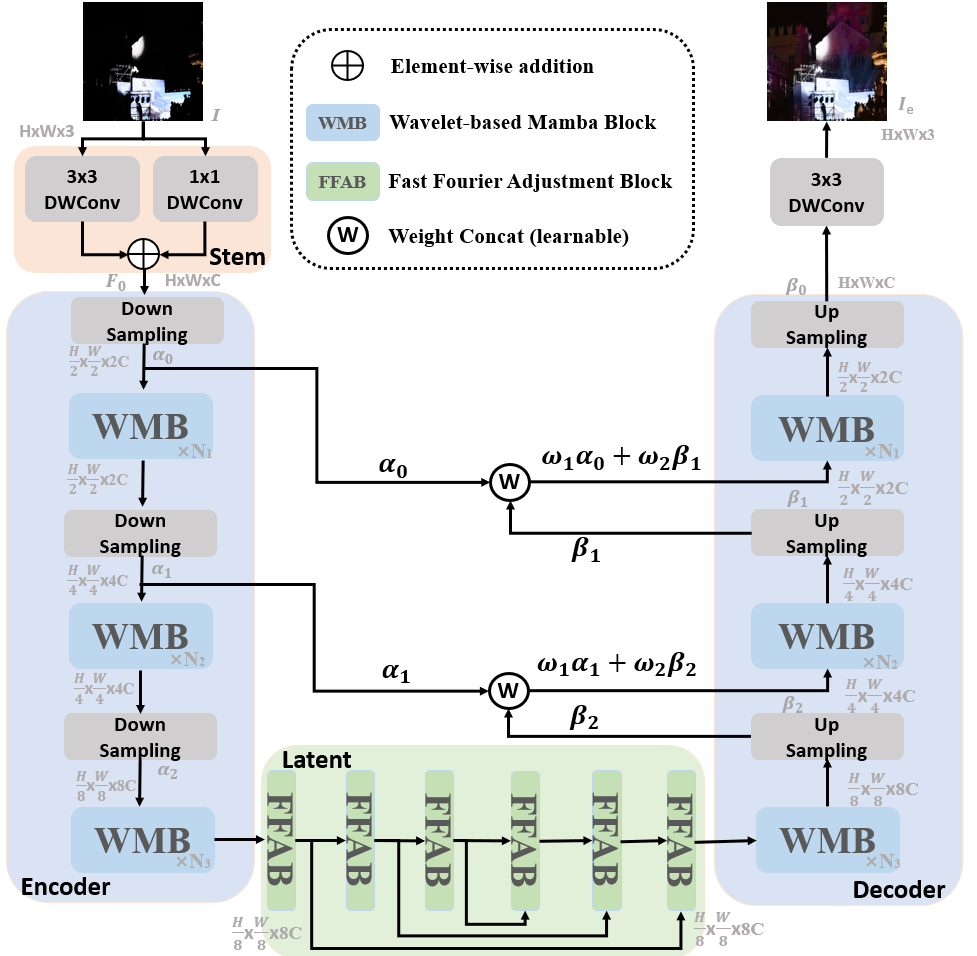}\\
\caption{The overview of WalMaFa architecture. Our model consists of an Encoder-Latent-Decoder structure that uses wavelet-based WMB to adjust global brightness during the Encoder and Decoder, and Fourier-based FFAB to adjust local details during the Latent.}\label{fig:3}
\end{figure}

\section{WalMaFa}
Common embedding-block structures in low-light image enhancement models include one-stage structure\cite{retinexformer}, multi-stage structure\cite{LLFormer} and Encoder-Decoder structure\cite{Restormer}. In order to leverage the strengths of the wavelet transform and the Fourier transform as analyzed in the motivation of Fig. \ref{fig:2}, we adopt an Encoder-Latent-Decoder structure. To be specific, we use a three-layer encoder and decoder composed of WMB to emphasize the multi-scale global brightness information, while a latent stage composed of FFAB is employed to fine-tune the local texture details.

As shown in Fig. \ref{fig:3}, the overview of WalMaFa is an Encoder-Latent-Decoder structure. We assume that the input low-light image is ${I \in {\mathbb{R}^{H\times W\times 3}}}$. ${I}$ is initially mapped to the intermediate dimension ${C}$ by the Stem, which consists of a two-way convolution (1${\times}$1 convolution and 3${\times}$3 convolution) to obtain ${F_0 \in \mathbb{R}^{H\times W\times C}}$ with double receptive fields. For encoder blocks, the hierarchical encoder consists of three layers of WMB and downsampling to achieve ${\alpha_{i}\in \mathbb{R}^{\frac{H}{2^i}\times\frac{W}{2^i}\times 2^iC}}$ and ${i = 0, 1, 2}$. For latent blocks, six FFABs stack consecutively and operate through sequential skip connections as follows:
\begin{equation}
\begin{aligned}
y_0 &= FFAB(\alpha_2),\\
y_i &= FFAB(y_{i-1}),i=\{1, 2, 3\},\\
y_4 &= FFAB(\mathbf{C}(y_1,y_3)),\\
y_5 &= FFAB(\mathbf{C}(y_0,y_4)),
\end{aligned}
\end{equation}
where ${\mathbf{C}}$ is channel concating operation. For decoder blocks, the hierarchical decoder consists of three layers of WMB and upsampling to achieve ${\beta_{i}\in \mathbb{R}^{\frac{H}{2^i}\times\frac{W}{2^i}\times 2^iC}}$ and ${i = 0, 1, 2}$. Besides, the ${\alpha_i}$ and ${\beta_{i+1}}$, ${i=0, 1}$ features are balanced for feature fusion through learnable weight control parameters ${\omega_1}$ and ${\omega_2}$. Finally, we use a 3${\times}$3 convolution to recover channels in RGB and achieve the estimated enhanced image ${I_e\in \mathbb{R}^{H\times W \times 3}}$.

 \subsection{Discrete Wavelet Transform and Fast Fourier Transform}

\noindent\textbf{Discrete Wavelet Transform.} Firstly, we briefly introduce the Haar Transform in 2D Discrete Wavelet Transforms (DWT). We use DWT to transform an input into four sub-bands, which contains colour-dominated low-frequency information and detail-dominated high-frequency information. Given a low-light feature map as input ${I\in \mathbb{R}^{H\times W \times C}}$, we employ Haar wavelets to decompose the input due to its simplicity and speed. The filters of the Haar wavelet transform are branched into low-pass ${L}$ and high-pass ${H}$, which can be expressed as:
\begin{equation}
\begin{aligned}
L = \frac{1}{\sqrt{2}}[1,1]^\top, H = \frac{1}{\sqrt{2}}[1,-1]^\top.
\end{aligned}
\end{equation}

After that, the input can be transformed into four sub-bands, which can be expressed as:
\begin{equation}
\begin{aligned}
I_{LL},\{I_{LH},I_{HL},I_{HH}\} = 2D-DWT(I),
\end{aligned}
\end{equation}where ${I_{LL},I_{LH},I_{HL},I_{HH}}$${\in}$ ${\mathbb{R}^{\frac{H}{2}\times \frac{W}{2} \times C}}$ represent color-dominated low-frequency component and detail-dominated high-frequency components in the vertical, horizontal, and diagonal directions, respectively. At a glance, it is equivalent to the downsampling operation in convolution. Downsampling in convolution adopts convolution kernels to compress the image size and expand the number of channels, resulting in the loss of some features, whereas DWT doesn't change the number of channels and doesn't lose information due to its bi-orthogonality property. Then, the IWT operation are used to reconstruct the output, which can be expressed as:
\begin{equation}
\begin{aligned}
I_{output} = 2D-IWT(I_{LL},I_{LH},I_{HL},I_{HH}).
\end{aligned}
\end{equation}

\noindent\textbf{Fast Fourier Transform}. Secondly, we briefly introduce the Fast Fourier Transform. We use Fast Fourier Transform (FFT) in Discrete Fourier transform due to its speed. Given a low-light input image ${x}$, whose shape is ${H \times W}$. We employ FFT to transform an input into two sub-bands, ${i.e.}$, color-dominated amplitude spectrum and detail-dominated phase spectrum. The transform function ${\mathcal{F}}$ which converts ${x}$ to the Fourier space ${X}$ can be expressed as:
\begin{equation}
\begin{aligned}
\mathcal{F}(x)(u,v) &= \frac{1}{\sqrt{HW}}\sum\limits_{h=0}^{H-1}\sum\limits_{w=0}^{W-1}x(h,w)e^{-j2\pi(\frac{hu}{H}+\frac{wv}{W})}, \\
&= X(u,v),
\end{aligned}
\end{equation}
where ${h,w}$ are the coordinates in the spatial space and ${u,v}$ are the coordinates in the Fourier space, ${j}$ is the imaginary unit, ${X(u,v)}$ can be expressed as:
\begin{equation}
\begin{aligned}
X(u,v) = R(X(u,v))+jI(X(u,v)),
\end{aligned}
\end{equation}
where ${R(X(u,v))}$ and ${I(X(u,v))}$ represent the real and imaginary units of ${X(u,v)}$, respectively.

We can obtain the amplitude component ${\mathcal{A}(X(u,v))}$ and phase component ${\mathcal{P}(X(u,v))}$ as follows:
\begin{equation}
\begin{aligned}
\mathcal{A}(X(u,v))=\sqrt{R^2(X(u,v))+I^2(X(u,v))},
\end{aligned}
\end{equation}

\begin{equation}
\begin{aligned}
\mathcal{P}(X(u,v))=arctan[\frac{I(X(u,v))}{R(X(u,v))}],
\end{aligned}
\end{equation}
where ${R(X(u,v))}$ and ${I(X(u,v))}$ can also be expressed as:
\begin{equation}
\begin{aligned}
R(X(u,v)) = \mathcal{A}(X(u,v)) \times cos(\mathcal{P}(X(u,v))),\\
I(X(u,v)) = \mathcal{A}(X(u,v)) \times sin(\mathcal{P}(X(u,v))).
\end{aligned}
\end{equation}

Note that essentially, both the DWT and the FFT decompose the input into two components where one is sensitive to colour brightness and the other is sensitive to texture detail. But according to Fig. \ref{fig:2}, we conclude that the low-frequency LL component of DWT has a significant improvement in colour brightness over the amplitude component of FFT, while the phase component of FFT has a significant improvement in detailed texture over the high-frequency LH, HL, HH of DWT in terms of SSIM map. Specifically, the SSIM metric of exchanging LL is 0.08dB higher than the SSIM metric of exchanging amplitude, and conversely, the SSIM metric of exchanging phase on top of exchanging amplitude component made up for the previous 0.08dB difference between LL and amplitude. So we propose to enhance global brightness with the low-frequency component of DWT and recover local texture detail and smoothness with the phase component of FFT. The detailed implementations are in Sec. \ref{sec:WMB} and Sec. \ref{sec:FFAB}.

\subsection{State Space Model(SSM)}
\label{sec:SSM}
SSMs are recently proposed models inspired by continuous state space models in control systems. SSMs are linear time-invariant systems that map the input stimulation ${x(t) \in \mathbb{R}^L}$ to the output response ${y(t) \in \mathbb{R}^L}$. Mathematically, SSMs can be formulated as linear ordinary differential equations (ODEs) as follows:
\begin{equation}
\begin{aligned}
h^{\prime}(t) = Ah(t)+Bx(t),\\
y(t) = Ch(t)+Dx(t),
\end{aligned}
\end{equation}
where ${h(t)\in \mathbb{R}^N}$ is a hidden state, N is state size, ${A\in \mathbb{R}^{N \times N}}$, ${B\in \mathbb{R}^{N}}$ and ${C\in \mathbb{R}^{N}}$ are the parameters for ${N}$, and ${D \in \mathbb{R}^1}$ is the skip connection operation.

On top of that, a discretization version was proposed for deep learning, which can convert the ODE into a discrete function and align the model with the sample rate of the underlying signal present in the input data ${x(t)\in \mathbb{R}^{L \times D}}$.

The ODE can be further discretized via the zeroth-order hold (ZOH), which incorporates a timescale parameter ${\Delta}$ to convert the continuous parameters ${A, B}$ into discrete parameters ${\overline{A}, \overline{B}}$, which can be expressed as:
\begin{equation}
\begin{aligned}
h^{\prime}(t) = \overline{A}h(t-1)+\overline{B}x(t),
\end{aligned}
\end{equation}
\begin{equation}
\begin{aligned}
y(t) = Ch(t)+Dx(t),
\end{aligned}
\end{equation}
 \begin{equation}
\begin{aligned}
\overline{A} = e^{\Delta A},
\end{aligned}
\end{equation}
\begin{equation}
\begin{aligned}
\overline{B} = (\Delta A)^{-1}(e^{\Delta A}-I)\Delta B,
\end{aligned}
\end{equation}
where ${\Delta} \in \mathbb{R}^D$ and ${B, C \in \mathbb{R}^{D \times N}}$.
To facilitate the extraction of low-frequency information from the wavelet transform, inspired by Mamba\cite{mamba}, we propose Channel-wise Mamba to explore the variability of colour and brightness in the channel dimension.

\subsection{Wavelet-based Mamba Block (WMB)}
\label{sec:WMB}
\begin{figure}[t]
\centering
\includegraphics[width=\linewidth]{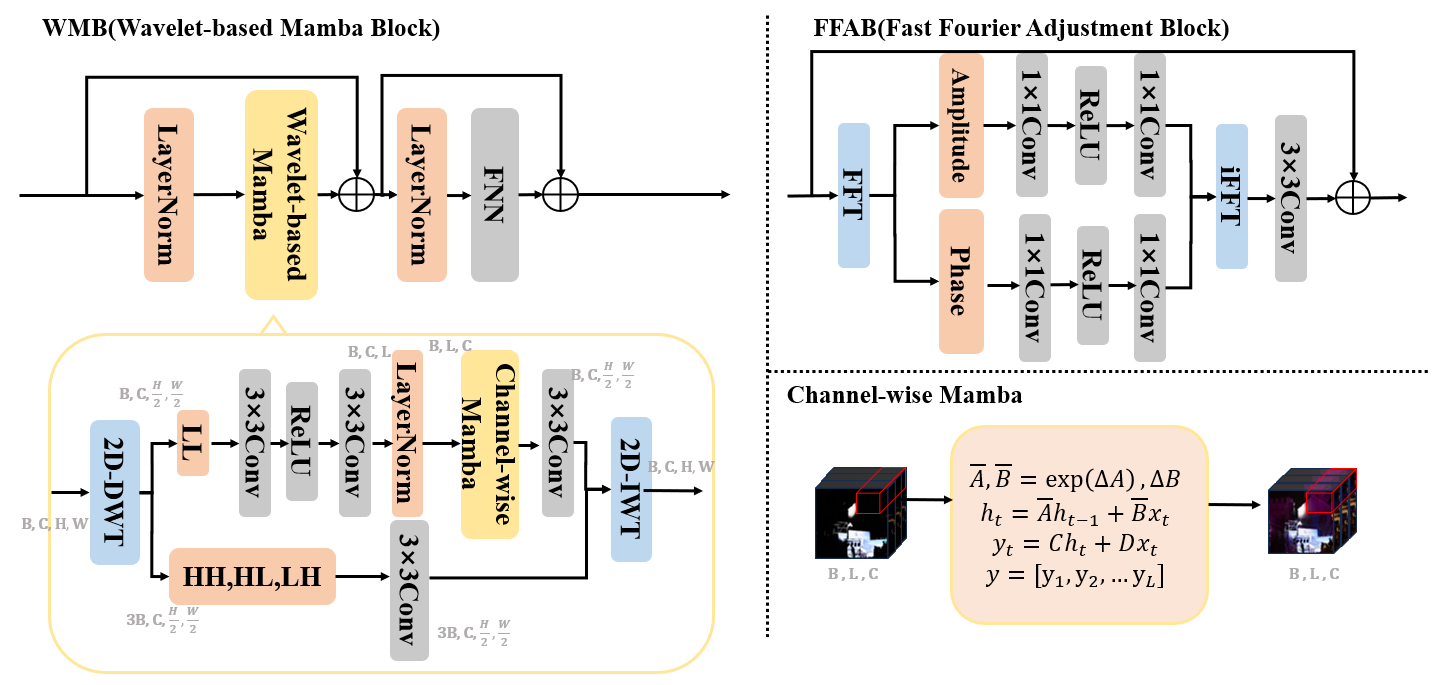}\\
\caption{The illustration of Wavelet-based Mamba Block (WMB) and Fast Fourier Adjustment Block (FFAB).}\label{fig:4}
\end{figure}
Based on the analysis in Fig. \ref{fig:2}, wavelet transform is more appropriate for extracting global colour brightness information due to its effective low-frequency component. So, we further introduce a Channel-wise Mamba\cite{mamba} module to complement spatial information where low-frequency information lacks in the channel dimension. We take advantage of Mamba's capability for linear analysis of long-distance sequences and offers fewer computational complexity compared to Transformer. The details of WMB are shown in Fig. \ref{fig:4}.

Given the input feature map ${x \in \mathbb{R}^{H \times W \times C}}$, WMB follows the efficient token mixer of Transformer, which can be expressed as:
\begin{equation}
\begin{aligned}
I^\prime &= WM(LN(x))+x,\\
I^{\prime\prime} &= FFN(LN(I^\prime))+ I^\prime,
\end{aligned}
\end{equation}
where WM denotes Wavelet-base Mamba operation and LN denotes LayerNorm operation.

\noindent\textbf{Wavelet-base Mamba.}
Given the input feature map ${F_{in} \in \mathbb{R}^{H \times W \times C}}$, ${F_{in}}$ is decomposed into four sub-bands: ${F_{LL}\in \mathbb{R}^{\frac{H}{2} \times \frac{W}{2} \times C}}$ and ${ \{F_{LH}, F_{HL}, F_{HH}\}\in}$ ${\mathbb{R}^{3 \times \frac{H}{2}\times \frac{W}{2} \times C}}$. Then, the block is split into two branches: low-frequency processing and high-frequency processing. For low-frequency processing, ${F_{LL}}$ is first fed into the 3${\times}$3 convolutional activation block and then merges the height and width dimensions to generate ${F_{LL}^C \in \mathbb{R}^{B,C,L}}$, where ${L=H \times W}$. This yields an integration of spatial features but retains the channels for subsequent modeling of the channel dimension. Following Channel-wise Mamba and 3${\times}$3 convolution, we obtain the low-frequency enhanced output ${F^\prime_{LL} \in \mathbb{R}^{\frac{H}{2}\times \frac{W}{2} \times C}}$ after a reshaping operation. For high-frequency processing, ${F_{LH}, F_{HL}, F_{HH}}$ are simply fed into the 3${\times}$3 convolution to generate ${\{F^\prime_{LH}, F^\prime_{HL}, F^\prime_{HH}\}\in \mathbb{R}^{3 \times \frac{H}{2}\times \frac{W}{2} \times C}}$. Finally, ${F^\prime_{LL}}$ and ${\{F^\prime_{LH}, F^\prime_{HL}, F^\prime_{HH}\}}$ are reconstructed to image space via an Inverse Wavelet Transform (IWT) operation to yield ${F_{out}\in \mathbb{R}^{H\times W\times C}}$.

\noindent\textbf{Loss Function}. Since the purpose of WMB is to focus on the global brightness of the image (${i.e.}$, low-frequency LL component). The loss involved in the wavelet-based mamba stage ${\mathcal{L}_w}$ is expressed as: 
\begin{equation}
\mathcal{L}_w = \Vert {F^\prime}_{LL}-G_{LL} \Vert_2,
\label{eq:ls_wmb}
\end{equation}
where ${{F^\prime}_{LL}}$ is low-frequency branch of the output, ${G_{LL}}$ is low-frequency branch of the ground-truth.

\subsection{Fast Fourier Adjustment Block (FFAB)}
\label{sec:FFAB}
Based on the analysis in Fig. \ref{fig:2}, Fourier transform is more appropriate for extracting local texture detail information due to its effective phase component. So we propose Fast Fourier Adjustment Block to facilitate the recovery of details. Given the input image ${F \in \mathbb{R}^{H \times W \times C}}$, we utilize the FFT to decompose ${F}$ into an amplitude component ${\mathcal{A}(F)}$ and a phase component ${\mathcal{P}{(F)}}$. Then, the two components are each fed into two 1${\times}$1 convolutional activation blocks to obtain ${\mathcal{A^\prime}(F)}$ and ${\mathcal{P^\prime}{(F)}}$. Finally, we reconstruct ${\mathcal{A^\prime}(F)}$ and ${\mathcal{P^\prime}{(F)}}$ to the image space ${F^\prime}$ via an inverse Fast Fourier Transform (iFFT) operation after a 3${\times}$3 convolution and skip connection.

\noindent\textbf{Loss Function}. Given the significant enhancement of texture detail by the phase component of the Fourier transform, the loss involved in the Fast Fourier Adjustment stage ${\mathcal{L}_f}$ is expressed as: 
\begin{equation}
\mathcal{L}_f = \Vert \mathcal{P}({F^\prime})- \mathcal{P}(G) \Vert_2,
\label{eq:ls_ffab}
\end{equation}
where ${{F^\prime}}$ is the output of the Latent, G is the ground-truth.

The overall Consistent loss, ${i.e.}$, the Charbonnier loss ${\mathcal{L}_c}$ is expressed as:
\begin{equation}
\mathcal{L}_c = \sqrt{\Vert {I_{e}}-G \Vert_2 +\epsilon^2},
\label{eq:ls_char}
\end{equation}
where ${I_{e}}$ is overall enhanced output of WalMaFa, ${\epsilon}$ is set as ${10^{-3}}$ empirically.

Finally, the overall loss ${\mathcal{L}_{total}}$ can be expressed as:
\begin{equation}
\mathcal{L}_{total} = \mathcal{L}_c + \lambda(\mathcal{L}_w + \mathcal{L}_f),
\label{eq:ls_overall}
\end{equation}
where ${\lambda}$ is a weight factor set as 0.1 empirically.

\section{Experiments}

\subsection{Environment and Datasets}
We evaluated our method on widely used LOL-v1\cite{Retinexnet} and LOL-v2\cite{LOL-v2} (${i.e.}$, LOL-v2-real, LOL-v2-synthetic) dataset. LOL-v1 dataset consists of 500 pairs of low-high images, of which 485 pairs are training and 15 pairs are testing. Most of images are indoor scenes. All images have a resolution of ${400\times600}$. The LOL-v2 dataset contains images from LOL-v1, and is split into v2-real and v2-syn. LOL-v2-real is captured in a real scene by varying ISO and exposure time with a resolution of ${400\times600}$. This subset includes 789 pairs of low/high images, of which 689 pairs are training and 100 pairs are testing. LOL-v2-synthetic synthesizes low-light images from RAW images by analyzing the illumination distribution of low-light images with a resolution of ${384\times384}$. This subset contains 900 pairs of low/high images for training and 100 pairs of low/high images for testing.

In addition to the above referenced paired datasets, we also introduced five unreferenced datasets in different shooting scenarios: DICM\cite{DICM}, LIME\cite{LIME}, NPE\cite{NPE}, MEF\cite{MEF} and VV\cite{VV} datasets that have no ground truth.

We trained WalMaFa on a server with three Tesla A10 GPUs with batch-size 12. The input images are cropped to ${128\times128}$. Adam\cite{2014Adam} optimizer was employed with 0.9 momentum for 5000 epochs. The learning rate was initially set to ${8\times10^{-4}}$ and decreased gradually by the cosine annealing scheme, reaching a minimum of ${1\times10^{-6}}$.

\subsection{Comparison with State-of-the-art Models}
\begin{figure}[tb]
\centering
\includegraphics[width=\linewidth]{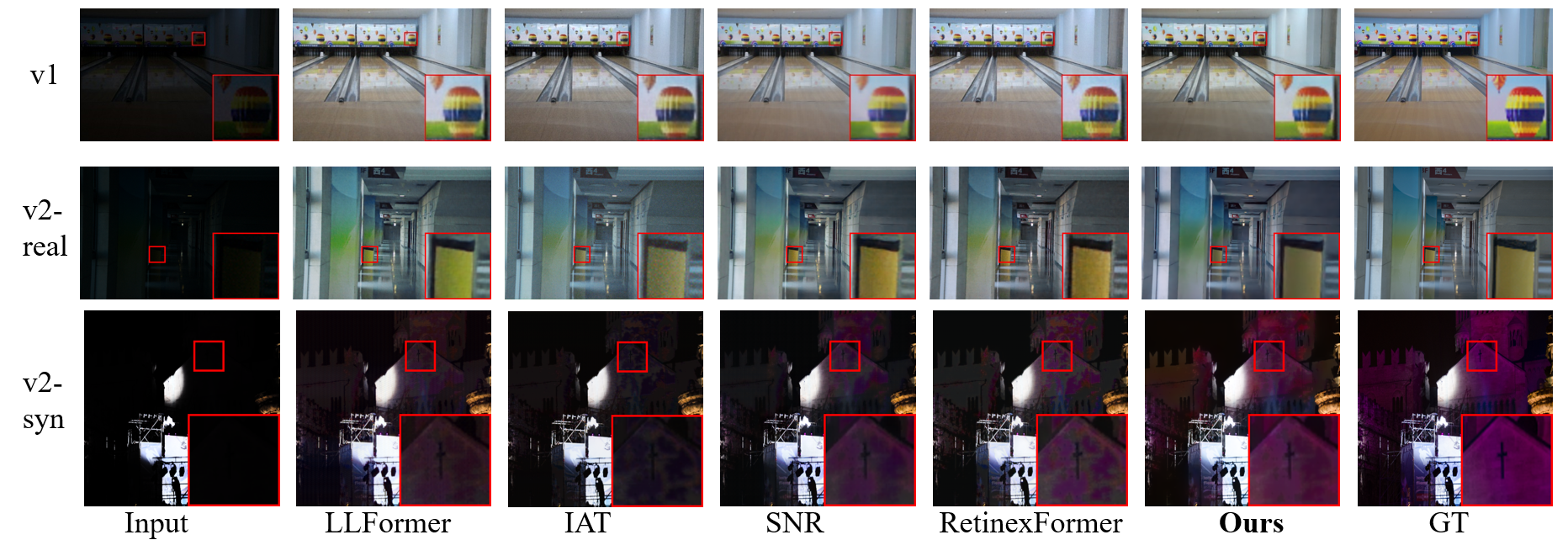}\\
\caption{The visual results of different models on LOL datasets.}\label{fig:5}
\end{figure}
\begin{figure}[tb] 
\centering
\includegraphics[width=\linewidth]{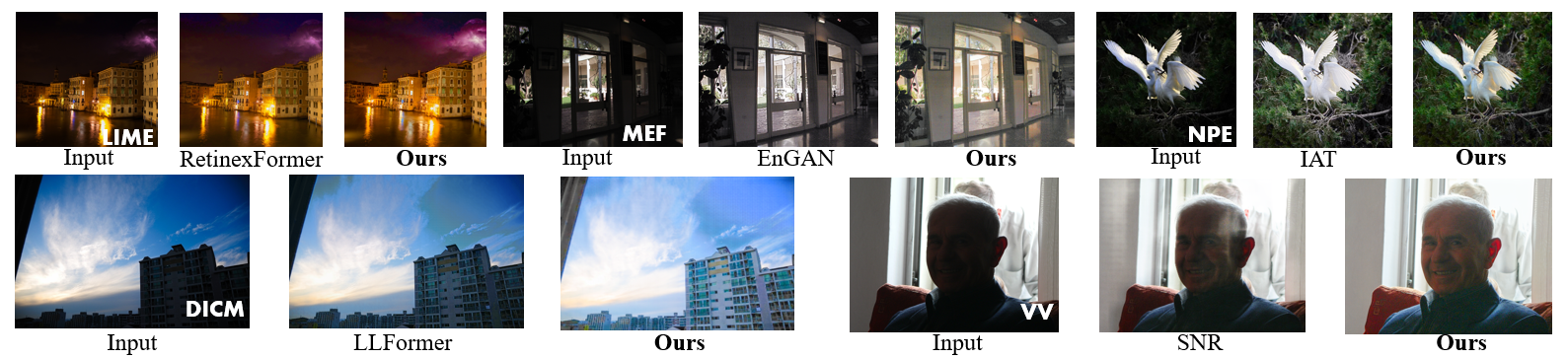}\\
\caption{The visual results of different models on VV, DICM, LIME, NPE, MEF datasets with LOL-v2-syn pretrained model.}\label{fig:6}
\end{figure}
\noindent\textbf{Low-light Image Enhancement Comparisons.} 
As shown in Table \ref{table1}, we selected the three most classic LOL datasets and tested the effect of 16 models on these datasets. In terms of Params(M)${\downarrow}$, FLOPs(G)${\downarrow}$ and Speed(s)${\downarrow}$ metrics, our model features a relatively lightweight modular design. In terms of PSNR/SSIM${\uparrow}$ metrics, our model achieves relatively excellent low-light enhancement performance. Note that when compared with the recent state-of-the-art model LLFormer\cite{LLFormer} on PSNR metric, WalMaFa outperforms LLFormer by 1.93 and 1.14 dB on LOLv2-real and LOLv2-syn datasets while saving 13.43M parameters and 22.63G FLOPs. When compared with UHD-Four, which is also based on Fourier transform, WalMaFa yields 0.18, 0.71, 1.31 dB improvements on the three benchmarks in Table \ref{table1}. The visual results are shown in Fig. \ref{fig:5}.

\noindent\textbf{Human Perception Study.} In order to make a more comprehensive analysis of the effects of low-light enhancement, we conducted a user perception score study with 15 participants to evaluate users' sensory perceptions of the enhanced low-light images on five non-reference benchmarks with a whole of 128 pictures. The testers were instructed to observe the results from: (i) Is the result over/underexposed? (ii) Are the colors vibrant and reasonable? (iii) Is the result ambiguous? (iv) Is the result free of noises? All models are trained on the LOL-v2-syn dataset for its rich colour representation and variety of environments. The final rating is shown in Table \ref{table2} using a Likert scale\cite{2010Likert} ranging from 1 (worst) to 5 (best). From Table \ref{table2}, our model achieves the highest scores on three datasets (LIME, NPE, VV) and the second-highest scores on two datasets (DICM, MEF). It is worth noting that our model also outperforms the second-place RetinexFormer\cite{retinexformer} by an average rating of almost 0.2. The visual results are shown in Fig. \ref{fig:6}.
\begin{table*}[t]
    \centering
    \caption{Quantitative comparisons on LOL datasets. The highest is in $\color{red}red$ color while the second highest is in $\color{blue} blue$ color.}   
	\resizebox{0.95\linewidth}{!}{
	\label{table1}        
	\begin{tabular}{l|c c|c c|c c|c c|c }  
		\toprule[1.5pt]   
		\multirow{2}{2cm}{Methods} & \multicolumn{2}{|c}{Complexity} &\multicolumn{2}{|c|}{LOL-v1} & \multicolumn{2}{|c|}{LOL-v2-real}&\multicolumn{2}{|c|} {LOL-v2-syn} &\multirow{2}{1.2cm}{Speed(s)$\downarrow$} \\ 
		&Params (M)$\downarrow$ &FLOPS (G)$\downarrow$& PSNR $\uparrow$  & SSIM $ \uparrow$  & PSNR $\uparrow$  & SSIM $\uparrow$ &  PSNR $\uparrow$  & SSIM $\uparrow$   \\ 
		\midrule[1pt]     
    		SID\cite{SID}	& 7.76 & 13.75 & 14.35  & 0.436 & 13.27  & 0.444 & 14.96  & 0.556 &- \\ 
    		DeepUPE\cite{DeepUPE}	& 1.02 &21.10	& 14.39  & 0.485  & 13.27 & 0.452 & 15.03 & 0.599 &- \\ 
            ZeroDCE\cite{ZeroDCE} & 0.09 & 2.53	& 14.89  & 0.555  & 14.12 & 0.512 & 14.93 & 0.531 & 0.002\\ 
            IPT\cite{IPT}     & 115.31 &6888  & 16.32  & 0.512 & 18.88 & 0.796 & 19.12  & 0.782 &1.554  \\ 
            UFormer\cite{Uformer} & 5.29 &12.00 & 16.56  & 0.799 & 18.69 & 0.786 & 19.89  & 0.852 & 0.268 \\ 
            RetinexNet\cite{Retinexnet} & 0.84 &587.47  & 16.88  & 0.556 & 15.25 & 0.574 & 17.17  & 0.780 &0.841  \\ 
            EnGAN\cite{EnGAN} & 114.35 &61.01  & 17.68  & 0.650 & 18.23 & 0.617 & 17.21  & 0.725 &2.254 \\ 
            RUAS\cite{RUAS} & 0.003 & 0.83 & 18.01  & 0.735 & 18.21 & 0.723 & 16.55  & 0.635 &0.003\\ 
    		KinD\cite{KinD} & 8.02 &34.99 & 20.86  & 0.790 & 18.67 & 0.752 & 15.22  & 0.542 &0.380 \\ 
            IAT\cite{IAT}  & 0.02 & 1.44 & 22.18   & 0.790 & 20.30 & 0.752 & 22.96  & 0.856 &0.004  \\ 
            Restormer\cite{Restormer}  & 26.13  &144.25    & 22.43 & 0.823 & 19.94 & 0.827  & 22.27 & 0.649 &0.114 \\ 
    		LLFormer\cite{LLFormer} & 24.52 & 37.04     & 23.64& 0.816  & 20.56 & 0.819 &  24.42  & 0.914 &0.077 \\ 
            SNR\cite{SNR}  & 4.01    &26.35  &${\color{red}24.21}$ &${\color{blue}0.841}$   & 21.48 & $\color{blue}0.843$ & 24.20 & $\color{blue} 0.927$ &0.032 \\ 
            UHDFour\cite{UHDFourICLR2023} & 17.53 & 8.26 & 23.09  & 0.837 & $\color{blue}21.78$ & 0.842 & 24.25 & 0.921 & 0.069 \\
            IGAWN\cite{IGAWN}& - & 15.39 & -  & - & 21.21 & 0.840 & 22.14 & 0.901 & - \\
            RetinexFormer\cite{retinexformer}  & 1.61   &15.57  & ${\color{blue}24.01}$ & 0.832   & 21.65 & 0.835 & ${\color{blue} 25.10}$ & 0.925 &0.014  \\ 
        \midrule[1pt]     
            \textbf{Ours}		& 11.09 &14.41  & 23.27  & ${\color{red}0.851}$ & ${\color{red}22.49}$ &${\color{red}0.869}$  & ${\color{red}25.56}$  & ${\color{red} 0.945}$ & 0.068 \\ 
		\bottomrule[1.5pt]   
	\end{tabular} }
\end{table*}

\begin{table}[t]
    \caption{User study scores on VV\cite{VV}, DICM\cite{DICM}, MEF\cite{MEF}, NPE\cite{NPE}, LIME\cite{LIME} datasets.}   
	\centering     
    \label{table2}        
    \resizebox{0.7\linewidth}{!}{
	\begin{tabular}{c|c|c|c|c|c|c}
        \toprule[1.5pt]
             Methods & DICM & MEF & LIME  & NPE & VV  & AVG\\
        \midrule[1pt] 
             EnGAN\cite{EnGAN} & 2.80 & 3.46 & 2.53 & 2.00 & 2.13 & 2.584 \\
             IAT\cite{IAT} & 2.66 & 3.33  & 3.00 & 3.13& 2.73 & 2.970  \\
             SNR\cite{SNR} & $\color{red}3.86$ & 3.46  & 3.33 & 3.20 & 3.26 & 3.422  \\
             LLFormer\cite{LLFormer} & 3.46 & 3.66  & 3.73 & $\color{blue}3.93$ &$ \color{blue}3.66$ & 3.688  \\
             RetinexFormer\cite{retinexformer} & 3.66 & $\color{red}4.00$  & $\color{blue}3.93$ & 3.73& 3.46 & $\color{blue} 3.756 $ \\
             \textbf{ours} & $\color{blue}3.80$ & $\color{blue}3.93$  & $\color{red}4.13$ & $\color{red}4.00$ & $\color{red}3.80$ & $\color{red}3.932$ \\
        \bottomrule[1.5pt]
    \end{tabular}}
\end{table}

\subsection{Ablation Study}

\begin{table}[t]
\makeatletter\def\@captype{table}
\caption{Structure ablation on LOL datasets.}   
	\centering     
    \label{table4}        
    \resizebox{0.8\linewidth}{!}{
	\begin{tabular}{c|c|c|c|c|c}
        \toprule[1.5pt]
             Model & LOLv1 & LOLv2-real  &  LOLv2-syn & Flops(G) & Speed(s) \\
        \midrule[1pt] 
             FFAB-WMB-FFAB & 20.84/0.784 & 20.75/0.796 & 22.47/0.888 & 15.79 & 0.077 \\
             only FFAB w/o WMB & 22.59/0.832 & 21.45/0.836 & 23.56/0.912 & 23.74 & 0.085\\
             only WMB w/o FFAB& 23.00/0.838 & 21.97/0.855 & 24.12/0.928 & 13.98 & 0.055 \\
        \midrule[1pt]
             WMB w/o Mamba & 22.10/0.808 & 20.88/0.785 & 23.25/0.915 & 4.18 & 0.032 \\
             WMB w/ SA & 23.12/0.837 & 21.79/0.788 & 25.38/0.935 & 32.23 & 0.111 \\
        \midrule[1pt]
             \textbf{Ours} & \textbf{23.27/0.851} & \textbf{22.49/0.869} & \textbf{25.56/0.945} & 14.41 & 0.068 \\
        \bottomrule[1.5pt]
    \end{tabular}}
\end{table}
\begin{figure*}[t] 
\centering
\includegraphics[width=\linewidth]{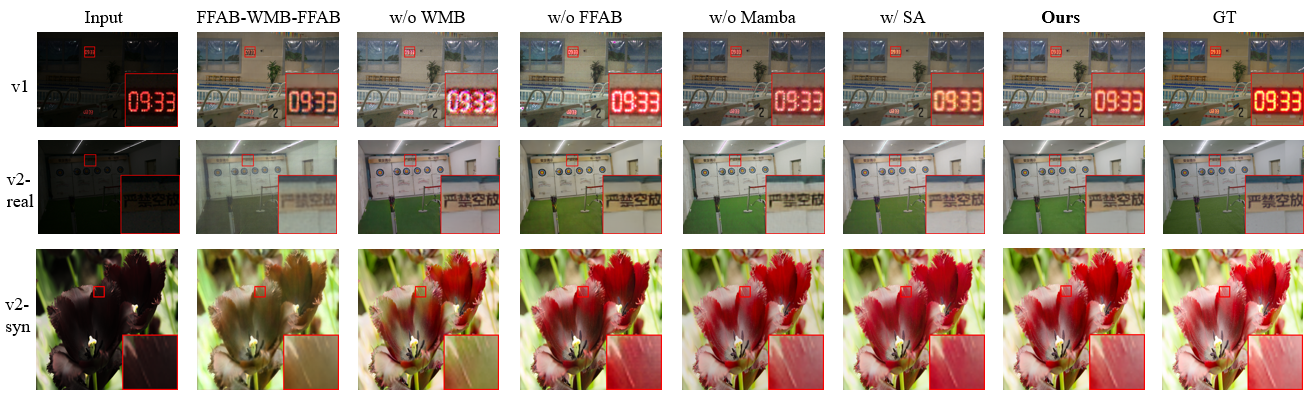}\\
\caption{The visual results of different settings of structure ablation on LOL datasets.}\label{fig:7}
\end{figure*}

\noindent\textbf{Structure Ablation}. We adopt the modular design of the model to verify the validity in Table \ref{table4}. We consider three module settings by reordering modules and removing proposed components. 

$\bullet$ ``FFAB-WMB-FFAB'' swaps the order of the WMB and FFAB modules of our model.

$\bullet$ ``only FFAB w/o WMB'' replaces WMB modules with 6-layer FFAB latent modules, so the model has only FFAB modules.

$\bullet$ ``only WMB w/o FFAB'' replaces FFAB modules with WMB modules, so the model has only WMB modules.

We also analyse the role of Mamba to the model and the improvement of Mamba over the SA(Self-Attention).

$\bullet$ ``WMB w/o Mamba'' removes Mamba from the WMB module.

$\bullet$ ``WMB w/ SA'' replaces Mamba with SA in the WMB module.

Our approach follows a global-local-global scheme, while the approach of ``FFAB-WMB-FFAB'' follows a local-global-local scheme. The approach of ``only FFAB w/o WMB'' and ``only WMB w/o FFAB'' enhance information in a single frequency domain (\ie, wavelet frequency domain or Fourier frequency domain). The results prove that both enhancing information in a single frequency domain and local-global-local scheme are inferior to the effectiveness of our approach. 

Table \ref{table4} also demonstrates that Mamba module is effective for low-light perception, offering superior performance and fewer computational complexity compared to the Self-Attention module. Fig. \ref{fig:7} shows the visual comparisons. Besides, we also conducted additional ablation studies in supplementary material.

\section{Conclusion}
In this work, we propose a novel Encoder-Latent-Decoder structure framework, namely WalMaFa. With brightness-dominated low-frequency components in the WMB and detail-dominated phase components in the FFAB, our model can ensure smoothness and clarity while enhancing the low-light image. In order to enhance global brightness, we also introduce channel-wise Mamba to extract low-frequency information in WMB. Comprehensive experiments prove the excellent performance, superior user-perceived effects, and fewer computing complexity of WalMaFa. 

Our future work involves exploring the feasibility of different components of the frequency transform method in other colour gamuts such as the HSV colour gamut.This time, we have only scanned Mamba in the channel dimension because of the channel’s effectiveness in enhancing brightness. In the future, we plan to scan Mamba in different directions.

\section*{Acknowledgements}
The authors would like to thank the anonymous reviewers for their invaluable comments. Any opinions, findings and conclusions expressed in this paper are those of the authors and do not necessarily reflect the views of the sponsors. This work was partially funded by the National Natural Science Foundation of China under Grant No. 61975124, State Key Laboratory of Computer Architecture (ICT, CAS) under Grant No.CARCHA202111, Engineering Research Center of Software/Hardware Co-design Technology and Application, Ministry of Education, East China Normal University under Grant No.OP202202, and Open Project of Key Laboratory of Ministry of Public Security for Road Traffic Safety under Grant No.2023ZDSYSKFKT04.

%
%
%
%

\title{Supplementary Materials}
\author{Junhao Tan$^{\dag}$\inst{1} \and
Songwen Pei$^{\dag*}$\inst{1} \and
Wei Qin\inst{1} \and
Bo Fu\inst{2} \and
Ximing Li\inst{3} \and
Libo Huang\inst{4}
{\\ \small{{$^{\dag}$}Contribute equally~~${*}$Corresponding author. Email address: swpei@usst.edu.cn}}}
\authorrunning{J. Tan et al.}
\institute{School of Optical-Electrical and Computer Engineering, University of Shanghai for
Science and Technology, Shanghai, 200093, China\\
\email{223330941@st.usst.edu.cn}, \email{swpei@usst.edu.cn}, \email{201440056@st.usst.edu.cn}
\and
School of computer science and artificial intelligence, Liaoning Normal University, Liaoning, 116081, China
\and
College of computer science and technology, Jilin University, Jilin, 134000, China
\and
School of Computer, National University of Dense Technology, Changsha, 410073, China
}
\maketitle
\section{Additional Ablation Studies}
\begin{table}[t]
\caption{Width and depth ablation on LOL-v1 dataset.}   
	\centering     
    \label{tables1}        
    \resizebox{0.5\linewidth}{!}{
	\begin{tabular}{c|c|c|c|c|c}
        \toprule[1.5pt]
             W & ${D_1}$ & ${D_2}$ & ${D_3}$   & Params (M) & PSNR/SSIM\\
        \midrule[1pt] 
             16 & 1 & 1  & 2 & 8.92 & 22.15/0.825 \\
             \rowcolor{lightgray}16 & 2 & 3  & 4 & 11.09 & \textbf{23.27/0.851}  \\
             16 & 4 & 4  & 4 & 12.49 & 22.60/0.831  \\
             16 & 4 & 6  & 8 & 20.16 & 22.99/0.850  \\
             32 & 2 & 3  & 4 & 41.86 & 22.12/0.842  \\
        \bottomrule[1.5pt]
    \end{tabular}}
\end{table}
\begin{figure}[t] 
\centering
\includegraphics[width=\linewidth]{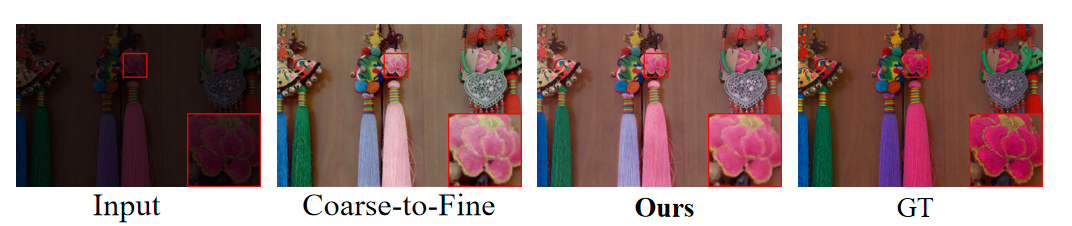}\\
\caption{The visual comparisons with coarse-to-fine pipeline.}\label{fig:s1}
\vspace{-5pt}
\end{figure}
\let\thefootnote\relax\footnotetext{Code is available at: 
\href{https://github.com/mcpaulgeorge/WalMaFa}{https://github.com/mcpaulgeorge/WalMaFa}}

\subsection{Width and Depth Ablation.}
 The width and depth of the model refer to embedding dimension and the number of iterations for each stage module, respectively. ${[D_1]}$, ${[D_2]}$, ${[D_3]}$ respectively indicates number of iterations for the WMB. The depth of [2, 3, 4] used for WalMaFa achieves the best performance as well as fewer parameters. 
 
\noindent\textbf{ Why larger models seem to perform worst?}
LOL-v1 dataset only consists of 485 train images and 15 test images, which inevitably leads to overfitting. Besides, we speculate that the deeper model will greatly overfit the global brightness due to ${D_1}$, ${D_2}$, ${D_3}$ indicating the number of iteration for WMB in the encoder-decoder, which will undermine the global and local balance.
\subsection{Why Encoder-Latent-Decoder?}
In this work, Encoder mainly aims at the coarse-grained global multi-scale brightness extraction (thanks to the low-frequency component of the WMB). Then, Latent fine-tines the fine-grained local details (thanks to the Phase component of the FFAB). However, we found that this coarse-to-fine pipeline exists a local overexposure problem (\ie, color distortion) caused by local texture smoothing, as shown in Fig. \ref{fig:s1}. So the extra coarse-grained Decoder is adopted to further balance the global brightness.
\begin{table}[H]
\caption{Structure ablation on LOL datasets.}   
	\centering     
    \label{tables2}        
    \resizebox{1\linewidth}{!}{
	\begin{tabular}{c|c|c|c|c}
        \toprule[1.5pt]
             Model & LOLv1 & LOLv2-real  &  LOLv2-syn & Flops(G) \\
        \midrule[1pt] 
             Unet & 21.18/0.833 & 20.80/0.821 & 23.18/0.898 & 11.94 \\
        \midrule[1pt] 
             Unet-skip-connection & 21.92/0.825 & 21.85/0.812 & 23.76/0.925 & 4.24 \\
        \midrule[1pt] 
             Channel-wise Self-Attention & 21.71/0.832 & 22.02/0.851 & 24.61/0.927 & 6.52 \\
             Simplified Channel Attention& 22.16/0.843 & 22.32/0.863 & 25.02/0.935 & 5.39 \\
        \midrule[1pt]
             \textbf{Ours} & \textbf{23.27/0.851} & \textbf{22.49/0.869} & \textbf{25.56/0.945} & 14.41\\
        \bottomrule[1.5pt]
    \end{tabular}}
\end{table}
\subsection{Supplementary Structure Abaltion.}
As shown in Table \ref{tables2}, we have experimented the Unet (Encoder with WMB and Decoder with FFAB) to verify the efficiency of Encoder-Latent-Decoder. We replace SSM with Unet-skip-connection between Encoder and Decoder to verify the efficiency of SSM. We also replace Channel-wise Mamba with Channel-wise Self-Attention (Restormer \cite{Restormer}) and Simplified Channel Attention (NAFNet \cite{chu2022nafssr}) to verify the efficiency of Channel-wise Mamba.
\end{document}